\documentclass{article}
\usepackage{spconf,amsmath,graphicx}

\newif\ifreview
\reviewfalse

\newcommand{\argmax}{\mathop{\rm argmax}\limits}
\newcommand{\taggedsymbol}[1]{\text{\textless #1\textgreater}}

\title{SIMULTANEOUS SPEECH-TO-SPEECH TRANSLATION SYSTEM \\WITH NEURAL INCREMENTAL ASR, MT, AND TTS}
\ifreview
\name{BLIND}
\address{BLIND\\ \vspace{3mm}}
\else
\name{Katsuhito Sudoh, Takatomo Kano, Sashi Novitasari, Tomoya Yanagita, Sakriani Sakti, Satoshi Nakamura}
\address{Nara Institute of Science and Technology\\
8916-5 Takayamacho, Ikoma, 630-0192 Japan}
\fi

\begin{document}
%\ninept
%
\maketitle

\begin{abstract}
This paper presents a newly developed, simultaneous neural speech-to-speech translation system and its evaluation.
The system consists of three fully-incremental neural processing modules for automatic speech recognition (ASR), machine translation (MT), and text-to-speech synthesis (TTS).
We investigated its overall latency in the system's Ear-Voice Span and speaking latency along with module-level performance.
\end{abstract}
\vspace{-0.1cm}
\section{Introduction}
\vspace{-0.1cm}
Computer-assisted cross-lingual conversation by automatic speech-to-speech translation has been one of the most challenging problems in spoken language technologies in decades \cite{sumita-etal-2007-nict,10.1145/2287710.2287712}.
Recent remarkable advances in speech and language processing led by deep learning techniques benefit this challenge by real-time and accurate speech translation.
%\memo{(Some mentions on non-incremental speech-to-speech translation studies by both cascade and end-to-end approaches?)} \cite{DBLP:conf/interspeech/JiaWBMJCW19} gogole multi-task model
One crucial problem in automatic speech-to-speech translation is its delay.
Spoken language processing tasks are usually handled in the utterance or sentence level.
Their application to the speech-to-speech translation suffers from a long delay that is proportional to the input length,
because the process starts after the observation of the end of an utterance.
That is similar to consecutive interpretation and is not useful for long monologues such as lecture talks.
On the other hand, in such situations, simultaneous interpretation is often used for an audience not proficient in the language of a talk.
Simultaneous interpretation is a challenging task to listen to the talk and speak its interpretation in a different language.

In this work, we tackle the problem of automatic simultaneous speech-to-speech translation and develop a neural system to do that from English to Japanese.
Here, we call our task simultaneous \emph{translation}, not simultaneous \emph{interpretation}.
We think the task of simultaneous interpretation includes some additional efforts for summarization to make the output concise for small latency and better understanding for the audience.
The problem requires real-time and incremental processing for the output generated simultaneously with the input.
Previous attempts to incremental neural speech translation focused on speech-to-text translation \cite{Niehues2018}.
Our work aims to speech-to-speech translation for natural information delivery by speech without a need for visual attention on text-based subtitles.
Our system is based on the cascade of three processing modules: incremental speech recognition (ISR), incremental machine translation (IMT), and text-to-speech synthesis (ITTS),
rather than recent end-to-end approaches
%\memo{(some references are needed here?)}
due to the difficulty of applying them to the simultaneous translation.

We follow existing studies on incremental neural speech processing.
For ASR, we choose an approach using a teacher-student training framework to train an incremental student model with the help of a non-incremental teacher model \cite{novitasari:2019:interspeech}.
For MT, we choose an approach called \emph{wait-k}, which delays the start of the decoding process simply by \emph{k} steps (i.e., \emph{k} input symbols from the ASR module) \cite{ma-etal-2019-stacl}.
For TTS, we choose approach starting the segmental speech synthesis after observing the next accent phrase \cite{yanagita:2019:ssw}.
These modules exchange their input/output symbols in the forms of subwords and work in a symbol-synchronous way,
so they can be cascaded even if they have different waiting strategies.

We also conduct a system-level evaluation of our system in system-level latency and module-level performance
on English-to-Japanese simultaneous translation on TED Talks.
The system-level latency measures are:
(1) processing delays for waiting and computation time, and (2) TTS speaking latency derived from overlaps of synthesized speech outputs.
The module-level performance is measured by standard metrics in ASR, MT, and TTS.
This work is the first attempt of system-level evaluation for a simultaneous speech-to-speech translation system and would be beneficial to future studies.

%The remainder of this paper is organized as follows.
%In section \ref{section:ssst}, we review the problem of simultaneous speech-to-speech translation, mainly in its difficulty.
%In section \ref{section:modules}, we describe the details of the incremental processing modules for ASR, MT, and TTS.
%In section \ref{section:evaluation}, we present system-wise evaluation of our system, followed by some discussions in section \ref{section:discussion}.
%We conclude this paper in section \ref{section:conclusion}.

\vspace{-0.1cm}
\section{Simultaneous Speech-to-Speech Translation}\label{section:ssst}
\vspace{-0.1cm}
The largest motivation of simultaneous speech translation is to convey spoken information in a different language with small latency.
For example, in a lecture talk,
a speaker often talks successively without any apparent pauses and sentence boundaries.
In such cases, a pause-based or sentence-based processing suffers from large latency until a pause or sentence boundary is observed.
The latency may become much larger in the latter part of a long talk due to accumulated processing delays.
Incremental processing helps catch up with the original speech.
Human simultaneous interpretation is demanded for the same reason.

Difficulty in simultaneous speech-to-speech translation is similar to that in simultaneous human interpretation.
In some language pairs with a distant word order such as English and Japanese,
human simultaneous interpretation is very challenging due to a large cognitive load to maintain long phrases in short-term memory in the interpreter's brain \cite{mizuno:2016}.

We show an example from the literature \cite{mizuno:2016}.
Suppose we are going to translate an English sentence:
\begin{quote}
    (1) The relief workers \underline{(2) say} (3) they don't have (4) enough food, water, shelter, and medical supplied (5) to deal with (6) the gigantic wave of refugees (7) who are ransacking the countryside (8) in search of the basics (9) to stay alive.
\end{quote}
Its usual translation into Japanese is:
\begin{quote}
    (1) \emph{kyuuen tan'tousha wa} (9) \emph{ikiru tame no} (8) \emph{shokuryo o motomete} (7) \emph{mura o arashi mawatte iru} (6) \emph{tairyo no nan-min tachi no} (5) \emph{sewa o suru tame no} (4) \emph{juubun na shokuryo ya mizu, shukuhaku shisetsu, iyakuhin ga} (3) \emph{nai to} \underline{(2) \emph{itte imasu.}}
\end{quote}
where the numbered phrases correspond each other.
Here we can identify the verb ``say'' is translated into ``\emph{itte imasu}'' at the end of the Japanese translation.
The following English phrases are translated in the reverse order.
This kind of translation causes a long delay and should be avoided for lowering interpreters' cognitive load and audience satisfaction.
Experienced interpreters try to translate such a different input by rephrasing (refer to the literature for details.)

While the cognitive load itself would not be a serious problem in a computer-based system,
we still have to wait for further input when we do not receive enough inputs to generate a partial translation.
This problem lies in the MT phase in the whole simultaneous speech-to-speech translation workflow.
ASR and TTS are regarded as sequential processes that can be solved in a left-to-right manner.
The use of broad context has some benefits in the results,
as revealed in previous studies on incremental ASR and TTS.
Thus, we employ incremental methods for all the processing modules and implement our simultaneous speech-to-speech translation system by a cascade of these modules.

Given an input sequence $X = x_{1}, x_{2}, ..., x_{|X|}$,
a general sequential transduction such as ASR, MT, and TTS predicts the corresponding output sequence $Y = y_{1}, y_{2}, ..., y_{|Y|}$.
In incremental processing, the predictions are made separately on subsequences.
Suppose an output subsequence $Y_{1}^{k} = y_{1}, y_{2}, ..., y_{k}$ has been predicted from partial observations of the input $X_{1}^{i} = x_{1}, x_{2}, ..., x_{i}$.
When we predict the next output subsequence $Y_{k+1}^{l} = y_{k+1}, ..., k_{l}$ after further partial observations $X_{i+1}^{j} = x_{i+1}, ..., x_{j}$,
the prediction is made based on the following formula:
\begin{equation}
    Y_{k+1}^{l} = \argmax_{\hat{Y}} P (\hat{Y} \mid X_{1}^{i}, X_{i+1}^{j}, Y_{1}^{k})
\end{equation}
where $\hat{Y}$ is a possible prediction of the subsequence.
We can use an input and output history in an arbitrary length by recurrent and self-attention neural networks and do not need further approximation with a strong independence assumption on a Markov process.
Note that the partial output $Y_{k+1}^{l}$ does not always corresponds directly to the observation $X_{i+1}^{j}$.
The partial output may cover just a portion of the partial observations $X_{i+1}^{j}$, include some delayed predictions on the past observations $X_{1}^{i}$, or contain aggressive predictions more than the given information by anticipations on future inputs.

One common problem here is how to decide the length of the partial observations to make partial predictions.
The most pessimistic way is to wait for the end of sentences, but we must shorten such a delay for \emph{simultaneous} translation.
In this work, we use fixed-length criteria for simplicity, as presented in the next section.
There have been some attempts to determine it adaptively according to the inputs in MT \cite{zheng-etal-2019-simpler,Chousa2019SimultaneousNM,zheng-etal-2020-simultaneous},
we reserve the problem as future work.

\vspace{-0.1cm}
\section{Incremental Processing Modules}\label{section:modules}
\subsection{Incremental Automatic Speech Recognition}
\vspace{-0.1cm}
\begin{figure}[t]
 \centering
 \centerline{\includegraphics[width=8.5cm]{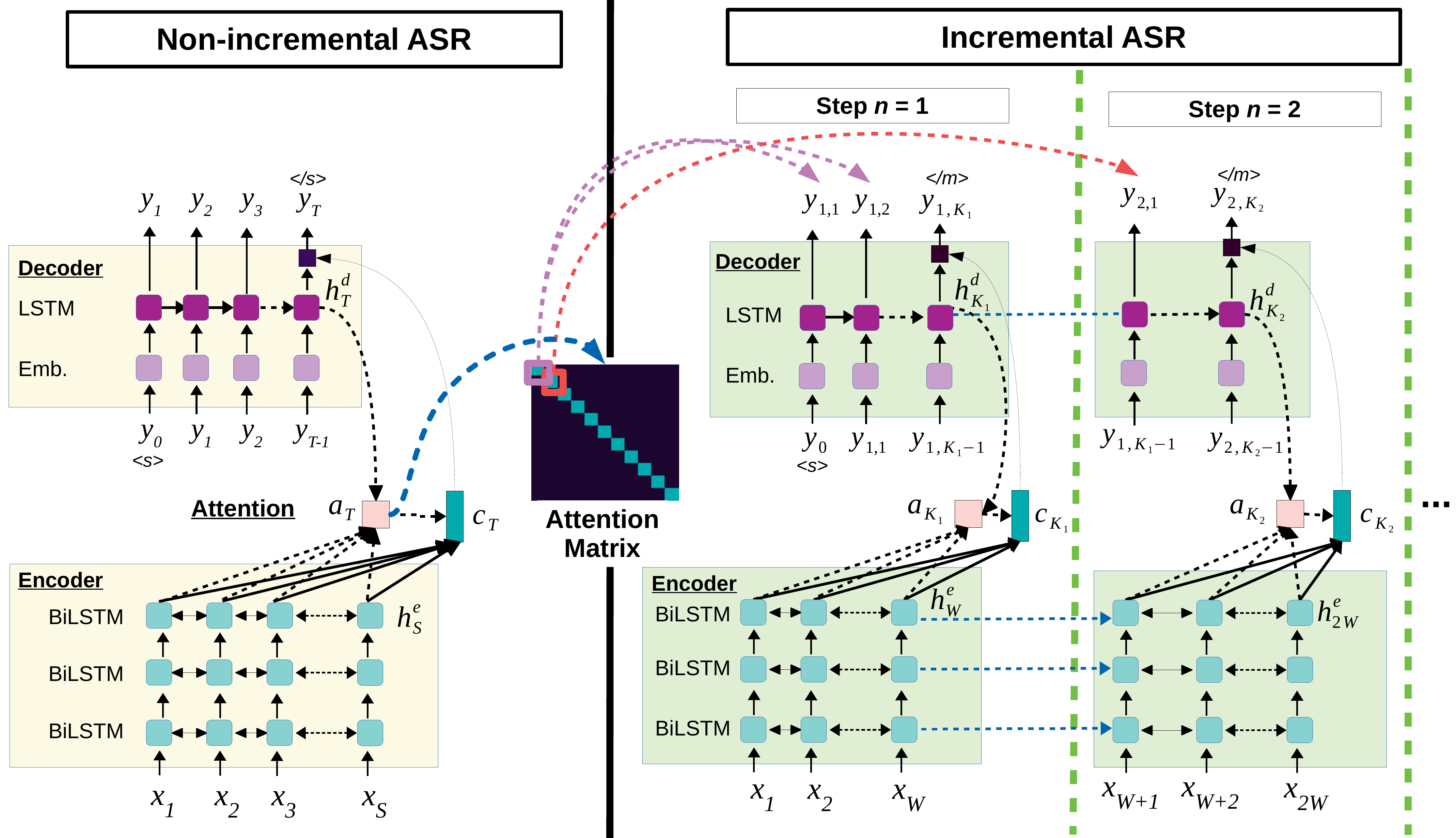}}
 \vspace{-0.2cm}
 \caption{Incremental Speech Recognition system. \cite{novitasari:2019:interspeech}}
\label{fig:ISR}
\end{figure}

An ASR system usually starts its recognition process after observing the end of a speech segment by voice activity detection.
When the speech segment is long,
the ASR transcription will be delayed together with its downstream processes of MT and TTS.
That causes serious latency in conversation aided by the speech-to-speech translation system.
To avoid this problem,
we implement an ISR following Novitasari et al.'s work \cite{novitasari:2019:interspeech}, as shown in Fig.\ref{fig:ISR}.
The ISR is trained with a teacher-student framework;
a non-incremental ASR is pre-trained as the teacher model,
and then the ISR is trained as the student model referring to the teacher model.
The student model learns to mimic the alignment between the input speech segments and the corresponding transcriptions given by the teacher model, through \emph{attention transfer}.

The teacher ASR model is used to transcribe an input speech segment $X=[x_1, \dots, x_I]$ ($I$: the number of time frames of the segment) into a token sequence of the source language $S=[s_0, \dots, s_N]$ ($s_0$ represents the beginning of a segment such as \taggedsymbol{s})
and also provides a sequence of attentions $A=[a_1, \dots, a_N]$, as follows:
\begin{eqnarray}
\textit{H} &=& \textit{Encoder} (X),\\ \nonumber
a_n,\hat{s}_n &=& \textit{Decoder} (\textit{H},s_{n-1}). \nonumber
\end{eqnarray}
Here, we assume that the encoder is a bidirectional RNN model
and that the decoder is a unidirectional, attentional RNN model.
$\textit{H}$ is a sequence of the encoder hidden states.
$\hat{s}_n$ is a predicted token, where $n$ is an index in the sequence.
We use the attention sequence $A$ to split the input speech $X$ into shorter sub-segments,
based on the assumption that the attention $a_n$ gives an alignment between $\hat{s}_n$ and its corresponding speech frames in $X$.
Note that $\hat{s}_n$ is the same as $s_n$ in the training phase.
%\memo{(Details of the segmentation should be described here.)}
As a result, the speech segment $X$ can be represented as a sequence of $M$ sub-segments, $\bar{X}=[\bar{x}_1, \dots, \bar{x}_M]$.
We put an end-of-block symbol \taggedsymbol{m} at the end of each sub-segment in the transcriptions $\bar{S}=[s_0, \bar{S}_1, \dots, \bar{S}_M]$.

The student ISR model can be used to transcribe the speech sub-segments $\bar{X}$ incrementally into the corresponding sub-segments of token sequences $\bar{S}$,
although it has the same architecture as the teacher ASR model.
The ISR method also uses \emph{look-ahead} decoding that predicts outputs looking at later frames.
The look-ahead decoding requires additional delays but improves recognition performance.

This ISR method achieved a 7.52\% in character error rate (CER) on a well-known WSJ dataset (WSJ SI-284 training set and eval92 test set) \cite{paul_1992_1}.
In comparison, another work that was based on unidirectional long short-term memory (LSTM) and connectionist temporal classification (CTC) \cite{Hwang2016} showed CER of 10.96\%.

\vspace{-0.1cm}
\subsection{Incremental Machine Translation}
\vspace{-0.1cm}

\begin{figure}[t]
 \centering
 \centerline{\includegraphics[width=8cm]{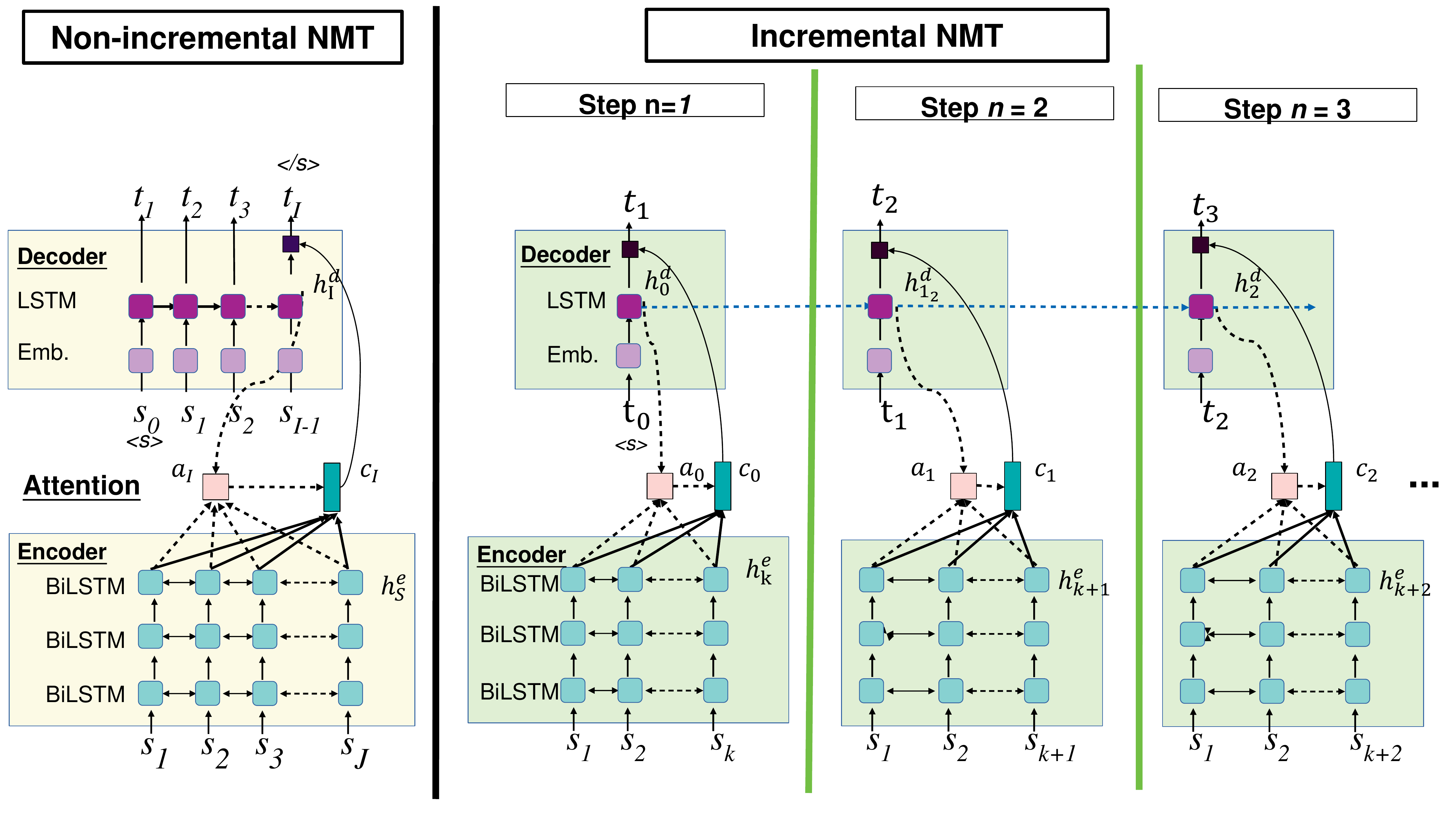}}
 \vspace{-0.2cm}
 \caption{Incremental Machine Translation system. \cite{ma-etal-2019-stacl}}
\label{fig:waitk}
\end{figure}

Machine translation has suffered from the problem of reordering,
due to the syntactic difference between the source and target languages.
The reordering problem is not so serious with current attention-based neural MT technologies \cite{Bahdanau2015NeuralMT,luong-etal-2015-effective,NIPS2017_7181}
but still very problematic when we consider incremental processing,
because we cannot observe full sentences.
Several previous studies tackled this problem by adaptive controls on reading source language inputs and writing target language outputs \cite{Fujita2013SimpleLC,gu-etal-2017-learning}.
In this work, we choose a recently-proposed approach called \emph{wait-k},
which delays the decoding process in $k$ input tokens \cite{ma-etal-2019-stacl} as shown in Fig. \ref{fig:waitk}, for simplicity.

The wait-k IMT model is used to translate a token sequence of the source language $S=[s_1, \dots, s_J]$ into that of the target language $T=[t_1, \dots, t_I]$ as follows.
\begin{eqnarray}
H_{i} &=& \textit{Encoder} (s_1, \dots, s_{i+k-1}),\\ \nonumber
\hat{t}_i &=& \textit{Decoder} (H_i, \hat{t}_1, \dots, \hat{t}_{i-1}). 
\end{eqnarray}
The decoder has to predict an output token based on the attention over an observed portion of the input tokens.
$k$ is a hyperparameter for the fixed delay in this model;
setting $k$ larger causes longer delays, while smaller $k$ would result in worse output predictions due to the poor context information.

This IMT method with $k=5$ achieved 21.53 in BLEU on a common ASPEC English-Japanese dataset \cite{nakazawa-etal-2016-aspec}.
Its non-incremental counterpart resulted in 32.22 in BLEU,
mainly due to the word order differences mentioned in section 2.

\subsection{Incremental Text-to-Speech Synthesis}
\begin{figure}[t]
 \centering
 \centerline{\includegraphics[width=5.5cm]{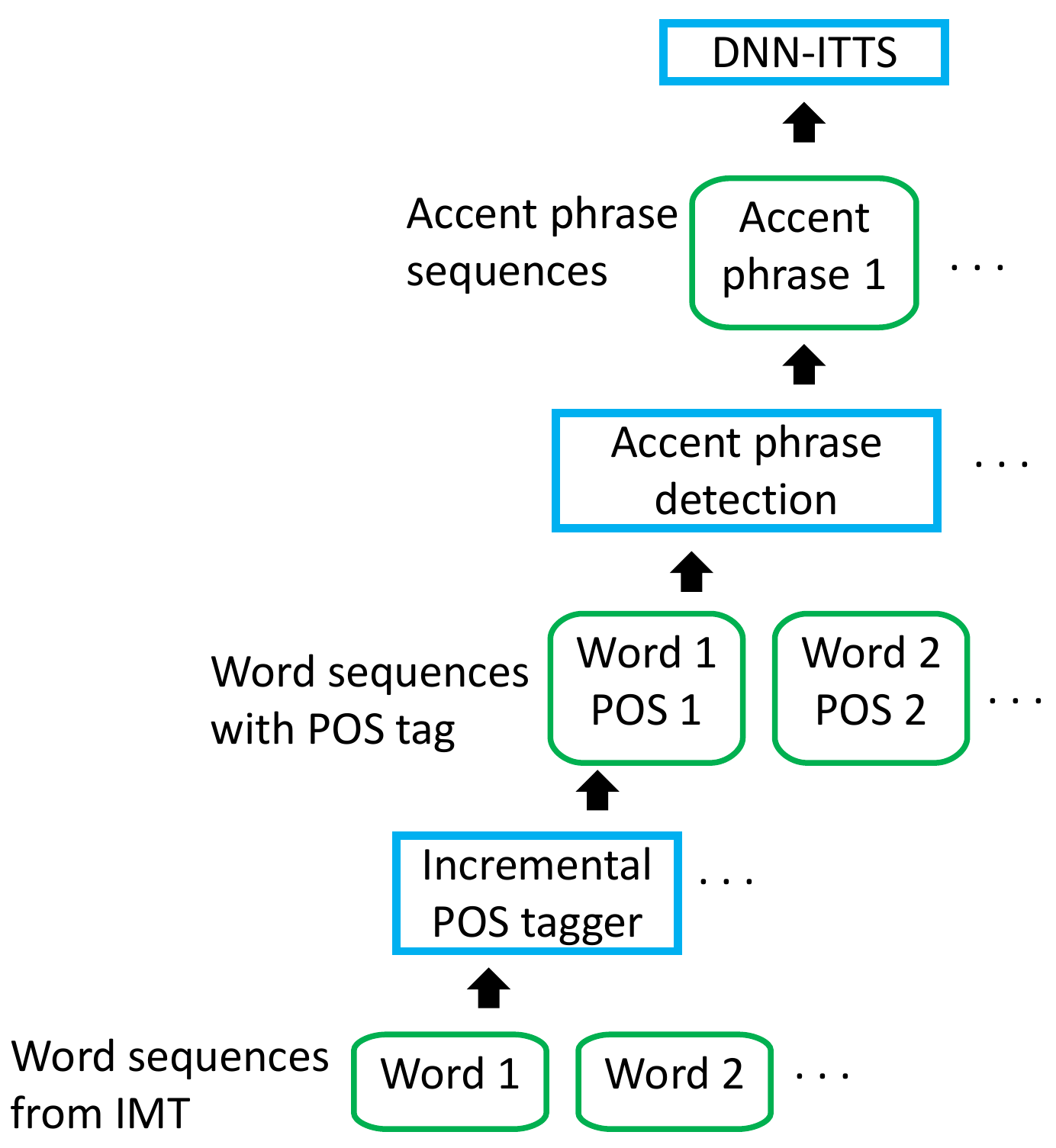}}
 \caption{Incremental text-to-speech synthesis system.}
\label{fig:iPOS}
\end{figure}

The workflow of our ITTS module is illustrated in Fig.~\ref{fig:iPOS}.
Since Japanese is a mora-timed language,
This module is based on a \emph{mora-timed} Japanese ITTS framework \cite{yanagita2018}.
A mora is a sub-unit syllable with one short vowel and any preceding onset consonant that corresponds to a single Japanese \emph{hiragana} phonogram \cite{suzuki2017accent}.
An \emph{accent phrase} is a mora sequence longer than a word.
The Japanese accent of each mora indicates the relative pitch change in the accent phrase and has an important role in the prosody,
similar to the tone type in tonal languages \cite{yokomizo2010}.

\begin{figure}[t]
 \centering
 \centerline{\includegraphics[width=6.5cm]{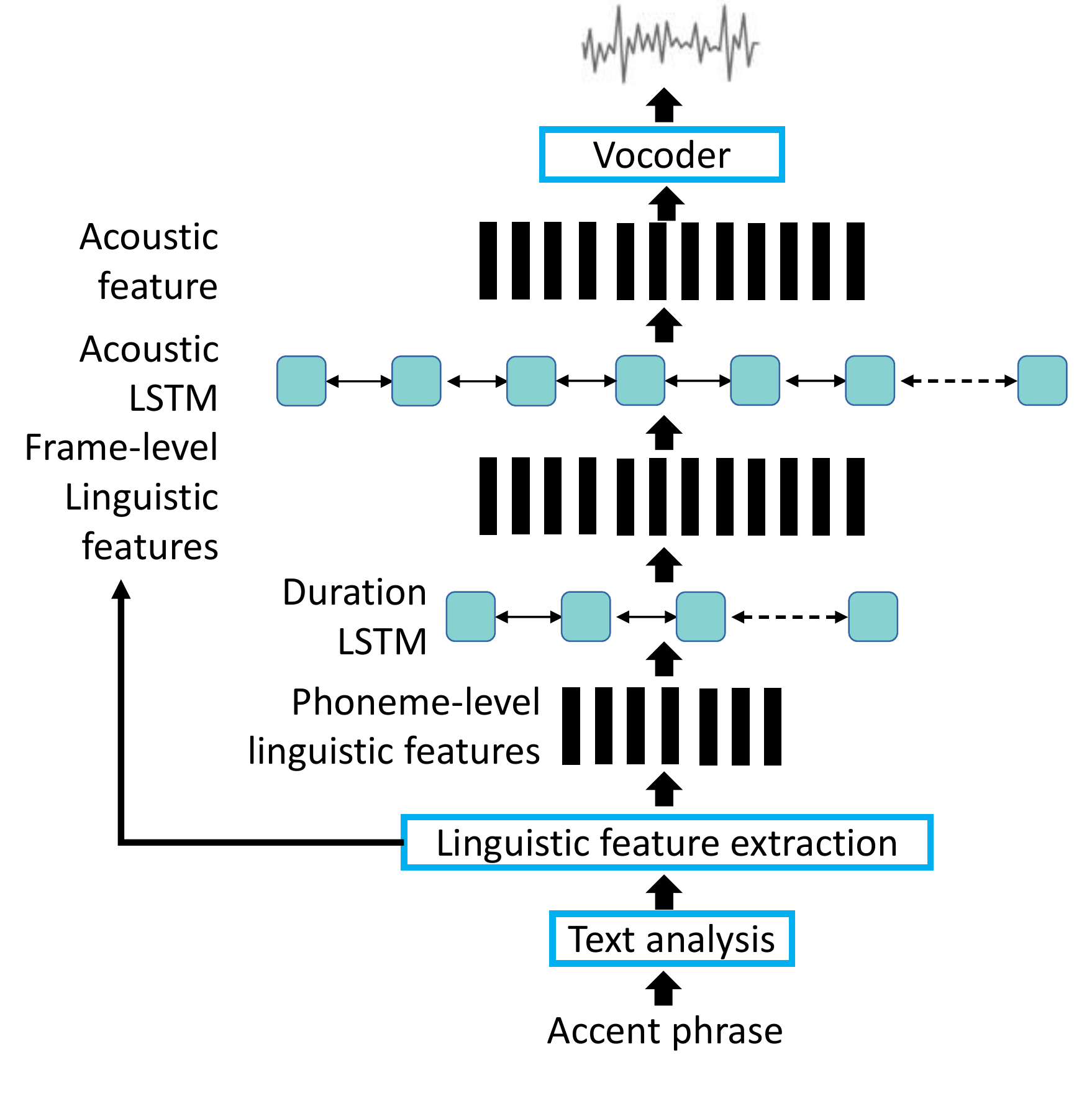}}
 \vspace{-0.1cm}
 \caption{DNN-based TTS system.  \cite{Zen:2015:icassp}}
\label{fig:tts}
\end{figure}

Yanagita et al. \cite{yanagita2018} synthesized speech outputs by accent phrases using linguistic features on the input text.
Our ITTS module is different from theirs as follows:
(1) we synthesize an output for a single accent phrase at a time without grouping it with future accent phrases;
(2) we exclude part-of-speech-based features in the synthesis step for simplicity;
(3) we choose a neural approach with LSTMs instead of HMMs.
Fig.~\ref{fig:tts} illustrates the ITTS
built upon a DNN TTS framework \cite{Zen:2015:icassp}.
It uses an accent phrase as a basis for generating a waveform through incremental part-of-speech tagging and rule-based re-segmentation on incremental inputs from the IMT (Fig \ref{fig:iPOS}).
It predicts the waveform for an accent phrase given by the preprocessing.
The waveform prediction consists of the following subprocesses,
as illustrated in Fig \ref{fig:tts}.
%the accent phrase sequence $\hat{T}={t_1, \dots, t_P}$, where $P$ is the number of accent phrases.
%In seconde phase, the ITTS resive the accent phrage and generates wavforme one by one as show in Fig \ref{fig:tts}.

\begin{enumerate}
    \item Linguistic features are extracted from the input accent phrase, in forms of a sequence of phonemes and accent information.
    Accent information consists of phoneme and accent type, their position in the accent phrase, and the number of moras in the accent phrase.
    \item Phoneme-level one-hot vectors are generated using the linguistic features.
    \item Phoneme durations are predicted using duration LSTMs.
    \item Frame-level vectors are generated from the one-hot vectors based on the phoneme durations. Positional information is concatenated with the frame-level vectors.
    \item Acoustic features are predicted using acoustic LSTMs and the frame-level vectors.
    \item A waveform is generated by the WORLD \cite{morise2016world} (D4C edition \cite{morise2016d4c}) vocoder from the acoustic features.
\end{enumerate}

\vspace{-0.1cm}
\section{Evaluation}\label{section:evaluation}
\vspace{-0.1cm}
We conducted a system-level evaluation of our system by an English-to-Japanese simultaneous speech-to-speech translation experiment.
The focuses in the evaluation were:
(1) \emph{System-level latency} and
(2) \emph{Module-level quality},
to investigate the detailed performance of the system in practice.

\vspace{-0.1cm}
\subsection{Evaluation Setup}
\vspace{-0.1cm}
We trained the ISR, IMT, and ITTS models with the settings shown in Table~\ref{tb:settings} and the following training procedures.

For ISR training, we used TED-LIUM release 1 \cite{DBLP:conf/lrec/RousseauDE12}.
We extracted $80$ dimension mel-spectrogram feature as an input.
The English speech transcriptions used in the training of ISR were segmented into subwords by SentencePiece \cite{kudo-2018-subword}.
The English SentencePiece model was shared with the IMT part.

The IMT model was LSTM-based attentional encoder-decoder model, trained using 1.41 million English-Japanese parallel sentences from WIT3 \cite{cettolo:2012:eamt} %(version 2017-01\footnote{\url{https://wit3.fbk.eu/mt.php?release=2017-01-trnted}} and ASPEC \cite{nakazawa-etal-2016-aspec} (only train1\footnote{\url{http://lotus.kuee.kyoto-u.ac.jp/ASPEC/}}).
The SentencePiece model was trained using the training sentences and the default training options other than the vocabulary size.
The subword vocabulary size was 8,000 for English and 32,000 for Japanese.

For ITTS model training, we utilized one female Japanese
speech dataset \cite{DBLP:journals/corr/abs-1711-00354} with 8k pairs of speech utterances and the corresponding transcriptions.
We used 5k utterances for the training set, 100 utterances for the development set, and 100 utterances for the test set.
Here we extracted 39 dimensions of Mel-generalized cepstral feature, 1-dimension of continuous log f0, 2-dimension of band averaged aperiodicity and 1-dimension of voiced/unvoiced flag as a target.
The frame period was set to 5ms.
We used a modified Open Jtalk\footnote{http://open-jtalk.sourceforge.net/} system for the text analysis system.
We trained an incremental part-of-speech tagger\footnote{https://github.com/odashi/incremental-tagger} using the Corpus of Spontaneous Japanese (CSJ) dataset \cite{DBLP:conf/lrec/MaekawaKFI00}. Phoneme duration was extracted forced alignment by HTS system\footnote{http://hts.sp.nitech.ac.jp/} and linguistic features.

The three modules were running in parallel and connected via UNIX pipes;
an output by a module was sent to \texttt{stdout}, and it was passed into \texttt{stdin} of the next module.
The system could be implemented in a server-client framework, but we used this na\"{i}ve implementation in this paper.

The test set consisted of two TED talks (Talk1 13:32, Talk2 17:19)
with English and Japanese subtitles and were not included in WIT3.
We could conduct a practical evaluation using a single input speech stream,
however, we used a pause-based segmented speech
because we observed significant speaking latency by TTS outputs as presented later in \ref{subsubsec:ttslatency}.
Note that the modules were implemented to flush its hidden vectors and be initialized when the end-of-sequence symbol \taggedsymbol{/s} was observed.

\begin{table}[t]
 \footnotesize
 \centering
 \caption{Model setting of the modules.}
 \label{tb:settings}
 \scalebox{1.1}[1.1]{
 \begin{tabular}{|l|c|}\hline
 \multicolumn{2}{|l|}{\textbf{ISR settings}}\\ \hline
 Input feature size & 80 \\ \hline
 Output vocabulary size & 8,009 \\ \hline
 Encoder input feed-forward sizes & 512 \\ \hline
 Decoder embedding size & 256 \\ \hline
 Encoder/Decoder RNN hidden size & 256 / 512 \\ \hline
 \# Encoder/Decoder RNN layers  & 3 / 1\\ \hline
 Attention function & Markov history MLP \cite{DBLP:conf/slt/TjandraS018}\\ \hline

\multicolumn{2}{|l|}{\textbf{IMT settings}}\\ \hline
 Input/Output vocabulary size & 8,000 / 32,000 \\ \hline
 \# Encoder/Decoder layers & 2 / 2\\ \hline
 RNN hidden size & 512 \\ \hline
 Delay "k" & 5 \\ \hline
\multicolumn{2}{|l|}{\textbf{ITTS settings}}\\ \hline
Duration / acoustic model input size & 297 / 300 \\ \hline
Duration / acoustic model output size & 1 / 44 \\ \hline
Hidden size & 512 \\ \hline
\# RNN layers & 1 \\ \hline
\end{tabular}

}
\end{table}
\subsection{System-level Latency}\label{subsec:latency}
We evaluated the latency of our system using two metrics: \emph{Ear-Voice Span (EVS)} and \emph{Speaking latency}.
\begin{figure*}[t]
 \centering
 \centerline{\includegraphics[width=16cm]{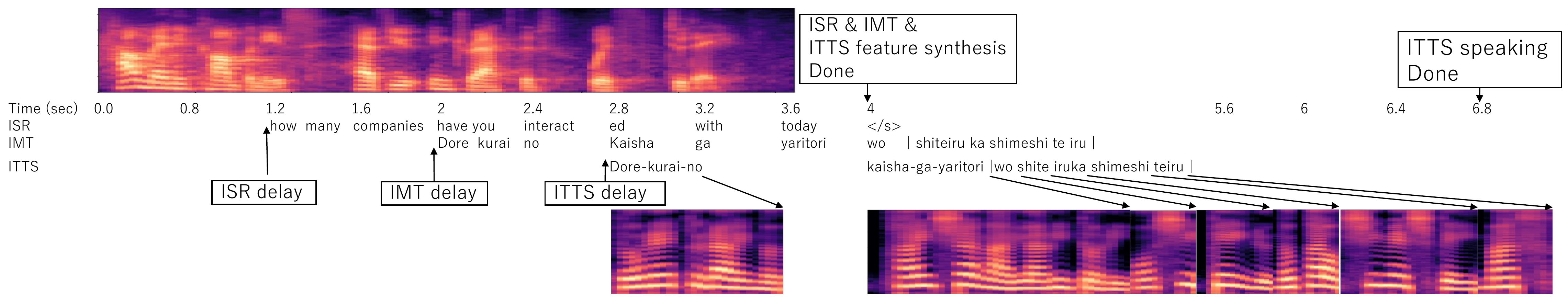}}
 \vspace{-0.2cm}
 \caption{ISR, IMT, and ITTS output and alignment example.}
\label{fig:sample}
\end{figure*}
\subsubsection{Ear-Voice Span (EVS)}\label{subsubsec:evs}
EVS is a common latency measure in the field of simultaneous interpretation.
EVS measures the delay between the start of a source speech and the start of the corresponding interpretation speech.
We measured the system-level delay in simultaneous speech-to-speech translation in a similar way for a human interpreter,
using the delay by the module-level processings and inter-module communications, which we call \emph{system-level EVS}. 

For the system-level EVS evaluation,
we aligned the ISR, IMT, and ITTS outputs with their output time information,
as shown in Fig.~\ref{fig:sample}.
The mel features were segmented in every 32 frames (0.55 seconds).
The ISR delayed three blocks, IMT delayed five blocks, and ITTS delayed seven blocks in this example.
Both of ISR and IMT delays are controlled by the hyper-parameters.
The ITTS delay depends on the output timing of the accent phrase generator.
We averaged module-level delays in the test set speech that was further segmented into 170 short streams.
The delays are shown in Table~\ref{tb:delay}.

\begin{table}[t]
 \footnotesize
 \centering
 \caption{Module-level processing delays from the beginnings of speech inputs by the ISR, IMT, and ITTS. Note again later modules have to work after their preceding module, so their delays include the delay in the preceding module.}
 \label{tb:delay}
 \scalebox{1.2}[1.2]{
 \begin{tabular}{|c|c|c|}\hline
 Module     & Mean        & Variance \\ \hline
 ISR delay  & 1.24 (sec) & 0.031 \\ \hline
 IMT delay  & 2.04 (sec) & 0.192 \\ \hline
 ITTS delay & 3.218 (sec) & 0.377\\ \hline
\end{tabular}

}
\end{table}
From Table~\ref{tb:delay}, we can see that the ITTS module brought long delays, and the delays were highly influenced by different inputs.
The delays and their variance in the IMT module were less than those by the ITTS, and the ISR module worked with almost fixed delays.

\vspace{-0.1cm}
\subsubsection{Speaking latency}\label{subsubsec:ttslatency}
\vspace{-0.1cm}

We found large delays derived from ITTS outputs in our system development.
A speaker of a TED-like talk speaks densely and does not put long pauses.
Thus, a new input came into the ITTS module even when the previous output waveform was being played, as shown in Fig.~\ref{fig:sample};
the latter parts of TTS outputs were queued and played with large delays.
This overlap would be a very serious problem in a long stream of the speech input.
So, we employed the segment-based evaluation in this work and considered the talk-level evaluation as future work.

\vspace{-0.1cm}
\subsection{Module-level Quality}
\vspace{-0.1cm}

We evaluated the module-level quality for all the three modules: ISR, IMT, and ITTS.
In the subjective evaluation described later,
we asked eleven people (graduate school students) to evaluate the results on a 1-5 scale.

\subsubsection{Incremental Speech Recognition}
We evaluated ISR results with word error rate (WER) and CER.
Table~\ref{tab:wercer} shows the results.
There were large differences between the two talks.

\begin{table}[t]
    \centering
    \caption{Evaluation results of Incremental Speech Recognition by word error rate (WER) and character error rate (CER).}
    \begin{tabular}{|c|r|r|}
        \hline
         Talk   & WER   & CER \\ \hline
         Talk 1 & 0.344 & 0.199 \\
         Talk 2 & 0.221 & 0.117 \\ \hline
    \end{tabular}
    
    \label{tab:wercer}
\end{table}

\vspace{-0.1cm}
\subsubsection{Incremental Machine Translation}
\vspace{-0.1cm}

We evaluated ISR+IMT results with BLEU-\{1,2,3,4\} \cite{papineni-etal-2002-bleu} using the reference from the corresponding voluntary translated Japanese subtitles.
Table~\ref{tab:bleu} shows the results by SacreBLEU\footnote{Fingerprint: BLEU+case.mixed+lang.en-ja+numrefs.1+smooth.exp+
tok.ja-mecab-0.996-IPA+version.1.4.13} \cite{post-2018-call}.
Unfortunately, the BLEU-4 score by the ISR+IMT translation was very low.
There would be various reasons, such as:
(1) Domain mismatch; the amount of in-domain training sentences was not large,
(2) Style mismatch; Japanese translations in TED Talks subtitles are usually not literal ones, so the surface-based evaluation by BLEU may not work well,
and (3) ASR error propagation.

\begin{table}[t]
    \centering
    \caption{Evaluation results of Incremental Machine Translation by BLEU-\{1,2,3,4\} and the length ratio between the hypothesis and referernce.}
    \begin{tabular}{|l|r|r|r|r|r|}
        \hline
         Talk & \multicolumn{4}{c|}{BLEU} & Length \\ \cline{2-5}
         & 1 & 2 & 3 & 4 & ratio \\ \hline
         Talk 1 & 26.2 & 12.9 & 6.6 & 3.5 & 1.161 \\
         Talk 2 & 29.6 & 14.8 & 8.2 & 4.9 & 1.127 \\ \hline
    \end{tabular}
    
    \label{tab:bleu}
\end{table}

We also conducted a subjective evaluation of the ISR+IMT results.
The subjective evaluation was based on 1-5 scale Adequacy and Fluency metrics used in past MT evaluation campaigns \cite{ldc2005}.
Table~\ref{tab:subjective_mt} shows the results.
As expected from the ISR and BLEU results,
the adequacy result on Talk 2 was slightly better than that on Talk 1.

\begin{table}[t]
    \centering
    \caption{Subjective evaluation results of Incremental Machine Translation by Adequacy and Fluency in a 1-5 scale. Average and standard deviation values are shown.}
    \begin{tabular}{|l|r|r|}
        \hline
         Talk & Adequacy & Fluency \\ \hline
         Talk 1 & 1.76 (0.86) & 2.54 (1.25) \\
         Talk 2 & 1.94 (0.87) & 2.48 (1.26) \\ \hline
    \end{tabular}
    \label{tab:subjective_mt}
\end{table}

\vspace{-0.1cm}
\subsubsection{Incremental Speech Synthesiss}
\vspace{-0.1cm}

Our ITTS evaluation was only with subjective evaluation,
because we do not have a reference speech in this task.
The subjective evaluation was based on Mean Opinion Score (MOS), focusing on the naturalness of the output speech rather than contents.
Table~\ref{tab:subjective_tts} shows the results.

\begin{table}[t]
    \centering
    \caption{Subjective evaluation results of Incremental Speech Synthesis by  Mean Opinion Score (MOS) in a 1-5 scale. Average and standard deviation values are shown.}
    \begin{tabular}{|l|r|}
        \hline
         Talk   & MOS \\ \hline
         Talk 1 & 2.34 (0.87) \\
         Talk 2 & 2.55 (1.01) \\ \hline
    \end{tabular}
    
    \label{tab:subjective_tts}
\end{table}

\vspace{-0.1cm}
\section{Discussion}\label{section:discussion}
\vspace{-0.1cm}

\begin{table}[t]
    \centering
    \caption{Translation examples.}
    \begin{tabular}{r|l}
        \hline
         En sub & here's another viewpoint of the world \\ \hline
         ISR & here's another point of the world \\ \hline
         Ja sub & \textit{mou hitotsuno sekai no mikata o} \\
                & \textit{ohanashi shimasu} \\ \hline
         IMT & \textit{kore wa betsu no shiten deno shiten desu} \\
             & (This is a viewpoint from another viewpoint.) \\ \hline\hline

         En sub & think about this dog here \\ \hline
         ISR & think about this dark here \\ \hline
         Ja sub & \textit{kono inu o rei ni tori masu} \\ \hline
         IMT & \textit{kono daaku enerugii o kangaete mite kudasai} \\
             & (Please think about this dark energy) \\ \hline
    \end{tabular}
    \label{tab:examples}
\end{table}

The latency results revealed
that our incremental system based on the cascade of three modules worked successfully with relatively small delays.
However, the quality results suggested the task difficulty
due to error propagation from ISR to IMT and lack of in-domain corpora in English-Japanese MT.
We show two translation examples in Table~\ref{tab:examples}.
The first one is a relatively good result, and the other one is a typical error propagation example.
Tight module integration would be promising, such as lattice-to-sequence \cite{sperber-etal-2017-neural}, but its extension to the simultaneous translation is not trivial.

Besides, we have no common evaluation metrics for simultaneous speech-to-speech translation other than module-level ones.
We used two latency metrics in this work, but the objective measurement of content delivery by speech-to-speech translation is crucial for further evaluation.

\vspace{-0.1cm}
\section{Conclusions}\label{section:conclusion}
\vspace{-0.1cm}

In this paper, we presented our English-to-Japanese simultaneous speech-to-speech translation system with its evaluation using English TED talks.
The system works fully-incremental with speech inputs,
by the cascaded modules of incremental ASR, incremental MT, and incremental TTS.
The latency evaluation revealed the module-level computation could be finished with about three seconds delay at maximum.
However, the system suffers from speaking latency.

Our future work includes further improvement of the modules in accuracy and efficiency, and controlling speaking duration to decrease the speaking latency.

\section{Acknowledgments}
Part of this work was supported by JSPS KAKENHI Grant Numbers JP17H06101.

% References should be produced using the bibtex program from suitable
% BiBTeX files (here: strings, refs, manuals). The IEEEbib.bst bibliography
% style file from IEEE produces unsorted bibliography list.
% -------------------------------------------------------------------------
\bibliographystyle{IEEEbib}
\bibliography{anthology,refs}

\end{document}